\documentclass[final,5p,times,twocolumn]{elsarticle}

\usepackage{amssymb}
\usepackage{amsmath}

\usepackage[utf8]{inputenc}
\usepackage[english]{babel}
\usepackage{graphicx}%
\usepackage{multirow}%
\usepackage{amsthm}%
\usepackage{amsmath,amssymb,amsfonts}
\usepackage{mathrsfs}%
\usepackage[title]{appendix}%
\usepackage[table, dvipsnames]{xcolor}
\usepackage{textcomp}%
\usepackage{manyfoot}%
\usepackage{booktabs}%

\usepackage{bbding}

\usepackage{hyperref}
\usepackage{cleveref}

\usepackage{graphicx}
\usepackage{textcomp}

\usepackage{threeparttable}

\usepackage{stackengine}
\newcommand\xrowht[2][0]{\addstackgap[.5\dimexpr#2\relax]{\vphantom{#1}}}
\usepackage{booktabs}
\usepackage{diagbox}

\usepackage{colortbl}
\usepackage{color}
\usepackage{xcolor}
\usepackage{fancyvrb}

\usepackage{algorithm}
\usepackage{algorithmic}

\usepackage{url}
\usepackage{caption}
\usepackage{subfig}
\usepackage{hyperref}
\usepackage{bbm}

\begin{document}

\begin{frontmatter}

\title{Bridge the Gap Between Visual and Linguistic Comprehension for \\Generalized Zero-shot Semantic Segmentation}  

\author[1,2]{Xiaoqing Guo\corref{cor1}}  

\author[3]{Wuyang Li\corref{cor1}}  

\author[3]{Yixuan Yuan\corref{cor2}} 

\cortext[cor1]{These authors contributed equally to this work. Emails are \\ \href{mailto:xiaoqing.guo@eng.ox.ac.uk}{xiaoqing.guo@eng.ox.ac.uk}, \href{mailto:wuyangli@cuhk.edu.hk}{wuyangli@cuhk.edu.hk}.}


\affiliation[1]{
    organization={Department of Computer Science, Hong Kong Baptist University}, 
    city={Hong Kong}, 
    country={China}
    }
    
\affiliation[2]{
    organization={Department of Engineering Science, University of Oxford}, 
    city={Oxford}, 
    country={United Kingdom}
    }

\affiliation[3]{
    organization={Department of Electronic Engineering, The Chinese University of Hong Kong}, 
    city={Hong Kong}, 
    country={China}
    }

\begin{abstract}
Generalized zero-shot semantic segmentation (GZS3) aims to achieve the human-level capability of segmenting not only seen classes but also novel class regions unseen in the training data through introducing the bridge of semantic representations, e.g., word vector. While effective, the way of utilizing one semantic representation to associate the corresponding class and to enable the knowledge transfer from seen to unseen classes is insufficient as well as incompatible with human cognition. Inspired by the observation that humans often use some `part' and `state' information to comprehend the seen objects and imagine unseen classes, we decouple each class into detailed descriptions, including object parts and states. Based on the decoupling formulation, we propose a Decoupled Vision-Language Matching (DeVLMatch) framework, composed of spatial-part (SPMatch) and channel-state (CSMatch) matching modules, for GZS3. In SPMatch, we comprehend objects with spatial part information from both visual and linguistic perspectives and perform graph matching to bridge the gap. In CSMatch, states of objects from the linguistic perspective are matched to compatible channel information from the visual perspective. By decoupling and matching objects across visual and linguistic comprehension, we can explicitly introspect the relationship between seen and unseen classes in fine-grained object part and state levels, thereby facilitating the knowledge transfer from seen to unseen classes in visual space. The proposed DeVLMatch framework surpasses the previous GZS3 methods on standard benchmarks, including PASCAL VOC, COCO-Stuff, and CATARACTS, demonstrating its effectiveness. 
\end{abstract}


\end{frontmatter}



\section{Introduction}
\label{sec:introduction}
Semantic segmentation aims to group pixels within an image into semantic classes. Though achieving great processes in real-world groundings, e.g., autonomous driving \cite{hoyer2022daformer} and medical interventions \cite{guo2022joint}, existing segmentation methods are still limited in generalizing to novel classes unseen in training data \cite{pourpanah2022review}. To break through this, by imitating human cognition, generalized zero-shot semantic segmentation (GZS3) has been proposed to identify unseen classes merely from some high-level semantic descriptions, including manually defined attributes \cite{chen2022transzero, chen2022msdn, zhao2022boosting} and word vectors \cite{baek2021exploiting, cheng2021sign, xu2022vgse}, without the access of unseen class data. 

\begin{figure}[t]
\centering
\includegraphics[width=88mm]{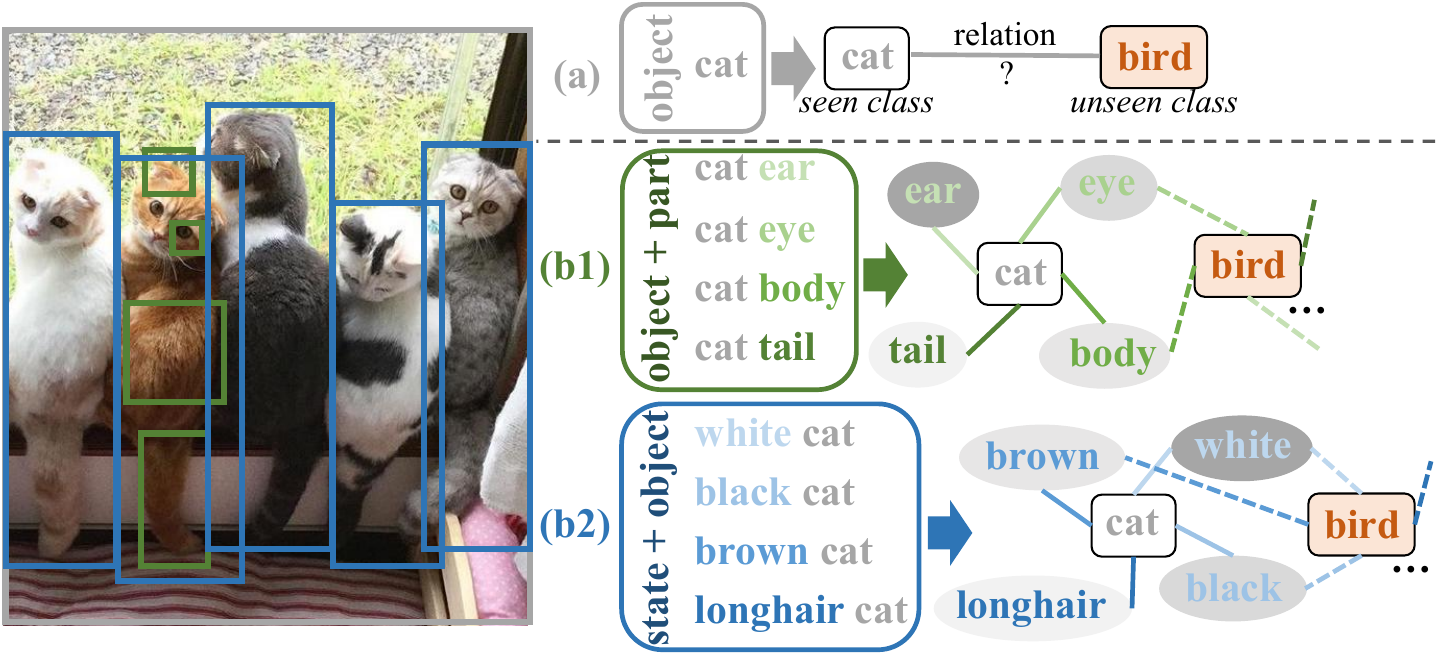}
\caption{Existing GZS3 methods usually use a single word vector or an attribute as semantic representation to associate each class, such as the `\textbf{\textcolor{gray}{cat}}' class (a). In contrast, inspired by the linguistic comprehension of human, we decouple the `\textbf{\textcolor{gray}{cat}}' class into detailed descriptions, including parts (b1) and states (b2) of the object, such as `\textbf{\textcolor{gray}{cat}} \textbf{\textcolor{LimeGreen}{eye}}' and `\textbf{\textcolor{CornflowerBlue}{brown}} \textbf{\textcolor{gray}{cat}}', which enables us to introspect the shared knowledge among classes and facilitate the generalizability of model on unseen classes, such as `\textbf{\textcolor{Bittersweet}{bird}}'.}
\label{fig:Motivation}
\end{figure}

Relying on the class-level semantic representation, existing GZS3 methods transfer knowledge from seen to unseen classes through either generative-based \cite{bucher2019zero, gu2020context, li2020consistent, shen2021conterfactual, cheng2021sign} or embedding-based \cite{baek2021exploiting, ding2022decoupling, xu2021simple, lu2021feature, zhang2021prototypical} strategy. The former learns a generative model to produce visual images or features for unseen classes based on seen class data and semantic representations of all classes \cite{cheng2021sign}. The latter learns an embedding space to associate the visual features of seen classes with their corresponding semantic representations and segments novel class regions by measuring the similarity between the visual and semantic representations in the embedding space \cite{zhang2021prototypical}. Despite the promising performance, existing GZS3 methods use a single and independent semantic representation, e.g., the word vector~\cite{bucher2019zero} and CLIP embedding~\cite{ding2022decoupling}, for each class (see Figure \ref{fig:Motivation} (a)), which has two limitations. (1) It is insufficient to capture adequate characteristics with a single semantic representation for each class, and the visual features under object occlusion are prone to deviate from the corresponding semantic representation accordingly. (2) In the linguistic aspect, the relationship among classes cannot be well discovered via a single semantic representation, preventing a reliable and explainable knowledge transfer from the seen to unseen classes. 

An intuitive observation is that, when \textit{comprehending} an object, we humans get the impression from both visual and linguistic perspectives. Instead of identifying one overview characteristic, humans usually describe an object by decoupling it into representative parts (e.g., `\textbf{\textcolor{gray}{cat}} \textbf{\textcolor{LimeGreen}{eye}}' in Figure \ref{fig:Motivation} (b1)) or states (e.g., `\textbf{\textcolor{CornflowerBlue}{brown}} \textbf{\textcolor{gray}{cat}}' in Figure \ref{fig:Motivation} (b2)) in linguistic comprehension, and the visual comprehension lies in detailed object parsing, texture, color, etc. Even though some object parts are shaded, humans can still make the correct recognition decision. Moreover, given descriptions of unseen class objects, humans usually \textit{introspect and retrieve} seen class object characteristics, such as parts and states, and then identify the unseen class object through shared characteristics between seen and unseen classes. 

Inspired by the \textit{comprehension} and \textit{introspection} in human cognition, we focus on decoupling objects to represent each class from visual and linguistic perspectives comprehensively, breaking the barrier of the limited single semantic representation. 
Hence, we propose a Decoupled Vision-Language Matching (DeVLMatch) framework to first decouple each class into \textit{parts} and \textit{states} to establish visual and linguistic graphs, and then formulate the knowledge transfer via graph matching, i.e., finding matching between visual nodes with linguistic counterparts to bridge the comprehension gap. Specifically, DeVLMatch consists of the spatial-part (SPMatch) and channel-state (CSMatch) matching modules. In SPMatch, we comprehend objects with local spatial features (visual space) and decoupled part information (linguistic space). Then, we construct a spatial visual graph on seen classes and a part linguistic graph on all classes to model the relation among decoupled components at a fine-grained level. With the well-formulated visual and linguistic graphs, a spatial-part graph matching is deployed to align cross graphs, and thus the part linguistic graph can act as a bridge to transfer the visual knowledge from seen to unseen classes. In CSMatch, since channels in feature maps represent texture or color information \cite{vittayakorn2016automatic, liang2020training}, we bridge the gap between channel information (visual space) and compatible object states (linguistic space) through graph matching. With comprehensive decoupled object descriptions, we explicitly discover and introspect the relationship between seen and unseen classes. After bridging the gap between visual and linguistic comprehension, the linguistic descriptions of unseen classes can facilitate generalizability in visual space with adequate semantic-level relations. Our main contribution is threefold: 
\begin{itemize}
 \item Imitating human cognition, we propose a brand new paradigm, DeVLMatch, for GZS3. By decoupling and matching objects from visual and linguistic perspectives, the segmentation model is able to \textit{comprehend} seen class objects and explicitly \textit{introspect} inter-class relationships to enhance unseen class generalizability. 

 \item We achieve decoupling and matching strategies from both object part and state levels, and consequently design SPMatch and CSMatch modules. Inspired by human intuition, SPMatch aligns the established spatial visual graph with the part linguistic graph built by decoupled object parts, and CSMatch matches channel visual graph with state linguistic graph.  
  
 \item DeVLMatch achieves state-of-the-art GZS3 results on standard benchmarks. The ablation analysis demonstrates that decoupling class description is superior to the single semantic representation by a large margin.
\end{itemize}

\section{Related Work}
\subsection{Zero-shot Learning}
Deep learning-based methods have achieved significant progress owing to the end-to-end solution for jointing feature extraction and classification. 
However, these models usually require abundant annotated data and are limited in recognizing novel classes unseen in training data \cite{pourpanah2022review}. 
Consequently, zero-shot learning (ZSL) that mimics human cognition in identifying unseen classes has been developed for image classification \cite{xian2018feature, kampffmeyer2019rethinking, yue2021counterfactual, feng2022non}. 
To achieve this, ZSL methods introduce semantic information, such as manually defined attributes \cite{chen2022transzero, chen2022msdn, chen2021semantics, zhao2022boosting} and word vectors \cite{baek2021exploiting, cheng2021sign, lu2021feature, xu2022vgse}, as a bridge to transfer the knowledge from seen to unseen classes in visual space. In conventional ZSL, the test set only contains unseen class data. Recently, a more realistic setting 
is generalized zero-shot learning (GZSL), where both seen and unseen classes are required to recognize in inference stage \cite{chao2016empirical, mercea2022audio, pourpanah2022review}. \cite{chao2016empirical} demonstrate that ZSL methods cannot perform well under GZSL setting due to the bias problem of classifying most unseen class data as seen classes, and then a calibrated stacking method is designed to penalize scores of seen classes. 
In this paper, we focus on more challenging and realistic GZSL task. 
By decoupling and matching objects, 
the model can mine shared knowledge among classes and introspect seen classes to enhance unseen class generalizability.

\subsection{Generalized Zero-shot Semantic Segmentation} 
Recently, increasing attempts have been made to extend GZSL to the semantic segmentation task, called generalized zero-shot semantic segmentation (GZS3). GZS3 targets to segment not only seen classes but also novel class regions. There are two main streams of GZS3 work: \textit{generative-based} \cite{bucher2019zero, gu2020context, li2020consistent, shen2021conterfactual, cheng2021sign} and \textit{embedding-based} methods \cite{baek2021exploiting, ding2022decoupling, xu2021simple, lu2021feature, zhang2021prototypical, zhou2023zegclip, liu2023model, he2023primitive, zhang2024csl}. \textit{Generative-based methods} train a generative model on seen class data and all class semantic information, so as to produce unseen class images or features. With the generated unseen class samples, a GZS3 problem can be transferred into a conventional supervised learning problem. For instance, ZS3Net \cite{bucher2019zero} first trains a generative model to output unseen class features in pixel level with the corresponding word vector as input. Then, both seen class features and generated unseen class features are leveraged to optimize the classifier for GZS3. \cite{gu2020context} introduce a context-aware prior to generate diverse and context-aware features for unseen classes. \cite{li2020consistent} exploits structural relationships between seen and unseen classes to constrain the generation of unseen class features. The general idea of \textit{embedding-based methods} is to learn projection functions to form a shared embedding space, where the visual features and semantic information are associated accordingly. In the shared space, unseen class regions can be segmented by measuring the similarity between visual feature and semantic representations. For example, 
\cite{baek2021exploiting} learn a joint embedding space, where the encoded semantic representations act as centers for visual features of corresponding classes. 
\cite{ding2022decoupling} group pixels into several segments, and then the pre-trained large-scale vision-language model is leveraged to align visual and semantic representations at the segment level. {\cite{xu2023side} also models the semantic segmentation task as a region recognition problem, where mask prediction and recognition are explicitly designed to be CLIP-aware, operating in an end-to-end manner.} 
However, almost all existing GZS3 methods use a single semantic representation to associate the corresponding class, which is insufficient to capture all characteristics within a class and hard to mine transferable knowledge for unseen class recognition. Different from existing methods that only use a single semantic representation, to the best of our knowledge, we represent the first effort in GZS3 that decouples to comprehensively represent each class from visual and linguistic perspectives.

\subsection{Graph Matching}
Graph Matching aims to find node correspondences across two graphs 
\cite{yan2016short, haller2022comparative, gao2021deep, sakaridis2018semantic, jiang2022graph},  
which has attracted increasing attention in computer vision tasks, such as multi-object tracking \cite{he2021learnable}, domain adaptation \cite{li2022sigma} and point cloud registration \cite{fu2021robust}. \cite{he2021learnable} perform graph matching between the tracklet graph and the intra-frame detection graph for multiple object tracking. \cite{li2022sigma} adaptively align the class-conditional distributions to bridge the domain gap via graph matching for domain adaptive object detection. \cite{fu2021robust} leverage graph matching to coordinate two homogeneous 3D point cloud sets, thereby facilitating point cloud registration. 
We are the first to formulate GZS3 task as a graph matching problem to bridge the gap between visual and linguistic comprehension.

\section{Decoupled Vision-Language Matching (DeVLMatch)}

\begin{figure}[t!]
\centering
\includegraphics[width=88mm]{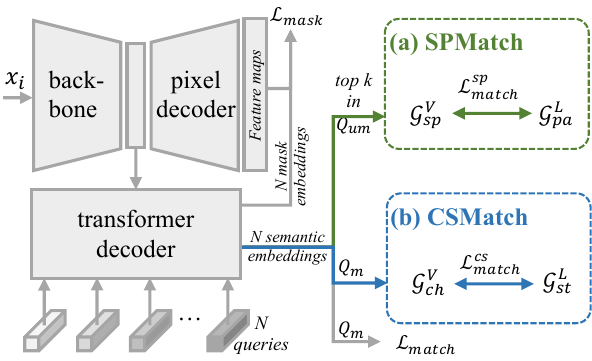}
\caption{Illustration of our Decoupled Vision-Language Matching (DeVLMatch) framework with (a) SPMatch and (b) CSMatch modules. The \textcolor{gray}{gray} part is baseline ZegFormer  \cite{ding2022decoupling}. }
\label{fig:Framework}
\end{figure}

\noindent \textbf{Problem description. }In this paper, we focus on the GZS3 task that generalizes the model to segment not only seen classes but also unseen class regions. Formally, we only have access to $N_s$ seen class images $\mathcal{X}_{train} = \{x_{i} \in \mathbb{R}^{H\times W\times 3}\}_{i=1}^{N_s}$ with spatial dimension of $H\times W$ and corresponding ground truths ${\mathcal{Y}}_{train} =\{{y}_{i}\in \mathbb{Z}^{H\times W} \}_{i=1}^{N_s}$ with $C_s$ seen classes. GZS3 aims to learn a segmentation model $\mathcal{F}(\cdot)$ that can correctly segment both seen and unseen class images $\mathcal{X}_{test} = \{x_{i} \in \mathbb{R}^{H\times W\times 3}\}_{i=N_s}^{N_{s'}} \cup \{x_{j} \in \mathbb{R}^{H\times W\times 3}\}_{j=1}^{N_u}$ in test stage. 
The number of unseen classes is $C_u$. Label spaces of seen and unseen classes are disjoint. 

Considering that manually defined attributes require labor-intensive annotations \cite{xu2022vgse}, we instead leverage text embedding to transfer the knowledge from the seen to unseen classes. In particular, each `class name' $c$ is inserted into a prompt template (e.g., `A photo of a \{class name\} in the scene') and then passed through a pre-trained text encoder \cite{radford2021learning} to obtain text embedding $t_c \in \mathbb{R}^{d}$. Hence, we have a text embedding set $\mathcal{T}=\{t_c \in \mathbb{R}^{d} \}_{c=0}^{C_s+C_u}$ containing both seen and unseen class semantic information. Note that $t_{0}$ is introduced to indicate the `no object' category \cite{ding2022decoupling}.  

\begin{figure*}[ht!]
\centering
\includegraphics[width=156mm]{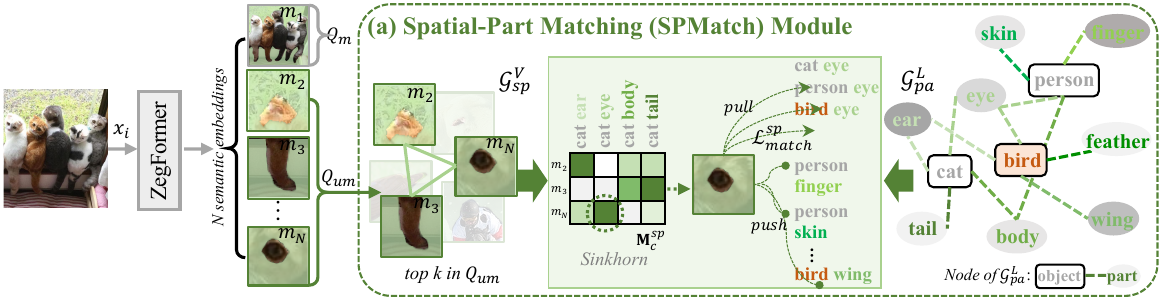}
\caption{Illustration of our spatial-part matching (SPMatch) module.}
\label{fig:F_SPMatch}
\end{figure*}

\noindent \textbf{Frameowrk overview. }The proposed Decoupled Vision-Language Matching (DeVLMatch) framework is illustrated in Figure \ref{fig:Framework}. We choose the state-of-the-art GZS3 framework, ZegFormer \cite{ding2022decoupling} as our baseline, and further advance spatial-part matching (SPMatch) and channel-state matching (CSMatch) modules to bridge the gap between visual and linguistic comprehension for GZS3. 

\noindent {\textbf{Baseline pipeline.} As shown in  Figure \ref{fig:Framework}, given an input image $x_{i}$, a feature map extracted from the backbone and $N$ segment queries are passed through a transformer decoder to mine correlations among queries and derive $N$ segment-level embeddings. Then, all segment embeddings are fed into a semantic projection layer and a mask projection layer to obtain the corresponding $N$ \textit{mask embeddings} and $N$ \textit{semantic embeddings} ($\{q_n \in \mathbb{R}^{d}\}_{n=1}^{N}$), respectively. {Note that the mask projection layer consists of a Multi-Layer Perceptron with 2 hidden layers, and the semantic projection is a linear layer.} 
The derived \textit{mask embeddings} are multiplied with the feature map from pixel decoder to generate \textit{class-agnostic segments}, which are supervised by the mask loss $\mathcal{L}_{mask}$. 
We use the combination of a dice loss \cite{milletari2016v} and a focal loss \cite{lin2017focal} as the mask loss. 
To further enable the recognition ability on segments, \textit{semantic embeddings} are constrained with $\mathcal{L}_{match}$ derived from ground truth classes, which are in the form of text embeddings to preserve semantic information. In particular, the predicted probability for a semantic embedding $q_n$ in ground truth class $c_{q}^{gt}$ is formulated as $p_n(c=c_{q}^{gt})=\frac{\exp \left(\frac{1}{\tau} s\left(t_c, q_n\right)\right)}{\sum_{c'=0}^{C_s} \exp \left(\frac{1}{\tau} s\left(t_{c'}, q_n\right)\right)}$, where $s(t,q)=\frac{t \cdot q^{\prime}}{|t|\left|q^{\prime}\right|}$ is the cosine similarity between two embeddings and temperature $\tau$ is set as 0.01 by default. Hence, the loss for constraining \textit{semantic embeddings} is $\mathcal{L}_{match}=-\log \left(p_n\left(c=c_q^{g t}\right)\right)$. According to the bipartite matching \cite{carion2020end, cheng2021per}, $\mathcal{L}_{match}$ will be calculated only if the segmentation prediction of the semantic embedding matches to a ground truth region; otherwise, no supervision will be generated. In the inference stage, the generated \textit{class-agnostic segments} with \textit{semantic embeddings} are used to compute segmentation predictions.}

\noindent {\textbf{DeVLMatch with SPMatch and CSMatch.}} Recognizing the insufficiency of using a single text embedding to associate the corresponding class in $\mathcal{L}_{match}$, we decouple each class in both visual and linguistic modalities and perform graph matching cross modalities to facilitate the generalizability on unseen class for GZS3. To this end, we design a spatial-part matching (SPMatch) module (Sec. \ref{spVLMatch}) and a channel-state matching (CSMatch) module (Sec. \ref{chVLMatch}) in DeVLMatch framework. In SPMatch, we comprehend classes with local spatial information in the visual graph and decoupled part information in the linguistic graph. Two graphs are matched by minimizing $\mathcal{L}_{match}^{sp}$ to mine shared object parts across different classes, thereby enhancing the unseen class recognition. In CSMatch, we bridge the gap between channel information in the established visual graph and compatible object states in the linguistic graph via $\mathcal{L}_{match}^{cs}$. Hence, our DeVLMatch framework is optimized by the joint loss of $\mathcal{L} =\mathcal{L}_{mask}+ \mathcal{L}_{match}+ \alpha \mathcal{L}_{match}^{sp} + \beta \mathcal{L}_{match}^{cs}$, where $\alpha$ and $\beta$ are weighting coefficients. 




\subsection{Spatial-Part Matching (SPMatch) Module}\label{spVLMatch}

\subsubsection{Spatial visual graph $\mathcal{G}_{sp}^{V}$}
Considering the single semantic representation is insufficient to capture the variant characteristics of object parts within an individual class, challenging the alignment between visual and semantic representations. Moreover, when an object is occluded, the extracted visual feature will deviate from the corresponding semantic representation due to the absent object parts. Therefore, it is desired to comprehend objects with decoupled parts, thereby explicitly aligning object parts across visual and linguistic perspectives. 

To decouple object parts in visual space, we derive the local visual features of an object from a spatial perspective, and construct a spatial visual graph $\mathcal{G}_{sp}^{V}$ (Figure \ref{fig:F_SPMatch}). Specifically, given $N$ decoded \textit{semantic embeddings} ${Q}$, a sub-set ${Q}_m \subset Q$ will be matched to ground-truth labels with bipartite matching \cite{carion2020end, cheng2021per}, and the rest unmatched queries ${Q}_{um}\subset Q$ represent background or partial foreground regions~\cite{cheng2021per} due to the one-to-one assignment constraint~\cite{carion2020end}. Considering that each \textit{semantic embedding} $q_n \in Q$ represents an independent local region $m_n$, we propose a simple yet effective solution to select informative queries to serve as graph nodes. We compute similarities between matched \textit{semantic embeddings} $Q_m$ and unmatched counterparts $Q_{um}$ within the training batch. Then top $k$ similar \textit{semantic embeddings} in ${Q}_{um}$ per class are selected to represent object parts and sent to a linear projection layer to obtain graph node embeddings. As for the graph edge, we leverage the semantic-level constraint \cite{jiang2022graph, zhang2021prototypical} by linking the queries within the same class, which formalizes a semantic space with structural awareness~\cite{li2022sigma}.

\begin{figure*}[ht!]
\centering
\includegraphics[width=156mm]{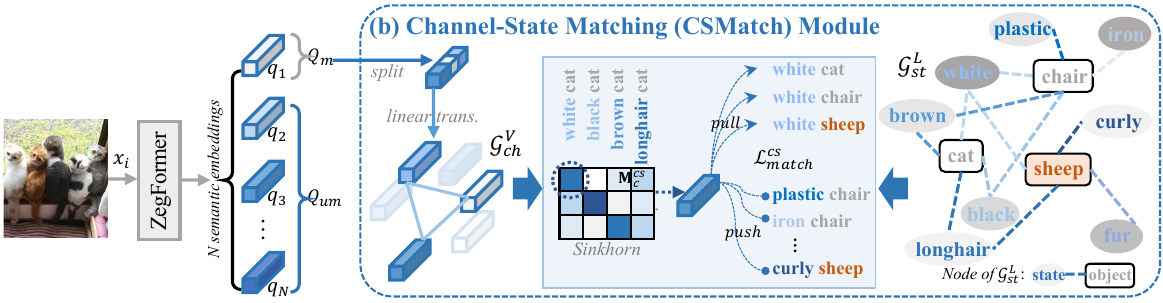}
\caption{Illustration of our channel-state matching (CSMatch) module.}
\label{fig:F_CSMatch}
\end{figure*}

\subsubsection{Part linguistic graph $\mathcal{G}_{pa}^{L}$}To build the part linguistic graph $\mathcal{G}_{pa}^{L}$ (Figure \ref{fig:F_SPMatch}), we take advantage of explicit relationships between nodes defined in ConceptNet \cite{speer2017conceptnet}, a commonsense knowledge graph. Specifically, ConceptNet defines 34 kinds of relations, such as `\textit{RelatedTo}', `\textit{IsA}', and `\textit{HasA}'. We can utilize the relation of `\textit{HasA}' to query object parts for each class. For instance, given a class `bird', we can find `legs', `wing', and `feather' are parts of a bird. 
Considering directly querying object parts would lead to very sparse $\mathcal{G}_{pa}^{L}$, we first construct a set of object parts $\mathcal{P}=\{P_1, P_2, ..., P_{M'}\}$ involving all classes. Then, we can establish a part linguistic graph $\mathcal{G}_{pa}^{L}$, where each node $t_{m, c}^{pa}$ is the text embedding of a phrase made by the object $O_{c} \in \mathcal{O}=\{O_1, O_2, ..., O_{C_s+C_u}\}$ and part $P_{m}$ pair, such as `bird wing', obtaining $M'\times(C_s+C_u)$ nodes. Note that the phrase is inserted into a prompt template and transferred into the text embedding to build the graph $\mathcal{G}_{pa}^{L}$. Then, we leverage ConceptNet to produce relational scores between object parts $\mathcal{P}$ and object classes $\mathcal{O}$. Since a large score indicates a stronger relevance, top $M$ object part pairs are selected for each class, where $M$ is empirically set as 10. Edges of $\mathcal{G}_{pa}^{L}$ among these selected object part pairs inner class or inner part are connected, otherwise unconnected.  

\subsubsection{Spatial-part graph matching}
With the spatial visual graph $\mathcal{G}_{sp}^{V}$ and part linguistic graph $\mathcal{G}_{pa}^{L}$, we then formulate the knowledge transfer in GZS3 as a graph matching problem and consequently devise the spatial-part graph matching strategy to narrow down the gap between visual and linguistic comprehension in spatial level. $\mathcal{G}_{sp}^{V}$ encodes relationships among seen class visual features, while $\mathcal{G}_{pa}^{L}$ embeds seen and unseen class semantic information. Aligned with $\mathcal{G}_{sp}^{V}$, the graph $\mathcal{G}_{pa}^{L}$ involving seen and unseen class relations can act as a bridge to enable the knowledge transfer from seen to unseen classes in visual space.

To this end, we perform spatial-part subgraph matching to associate each visual query node in $\mathcal{G}_{sp}^{V}$ to the linguistic part in $\mathcal{G}_{pa}^{L}$ without part-level labels, empowering the linguistic comprehension in visual recognition~\cite{ovd}. Specifically, for each class, according to the edge information in $\mathcal{G}_{sp}^{V}$ and $\mathcal{G}_{pa}^{L}$, connected nodes within the class form the subgraphs, denoted as $\hat{\mathcal{G}}_{sp}^{V}$ and $\hat{\mathcal{G}}_{pa}^{L}$. Taking class $c$ for example, we first compute a node affinity matrix $\mathbf{M}^{sp}_{c}$ cross subgraphs $\hat{\mathcal{G}}_{sp}^{V}$ and $\hat{\mathcal{G}}_{pa}^{L}$ to model the correspondence between visual and linguistic comprehension. Then, the affinity matrix $\mathbf{M}^{sp}_{c}$ is transformed to a sparse node permutation $\mathbf{\Pi}^{sp}_{c}$ through differential Sinkhorn iterations, establishing the subgraph matching between $\hat{\mathcal{G}}_{sp}^{V}$ and $\hat{\mathcal{G}}_{pa}^{L}$ as follows: 
\begin{equation}
\begin{small}
\begin{aligned}
\begin{split}
\mathbf{\Pi}^{sp}_{c}&=\max _{\mathbf{M}^{sp}_{c}} \operatorname{Tr} \left(\mathbf{M}^{sp \top}_{c} \hat{\mathcal{G}}_{sp}^{V \top} \hat{\mathcal{G}}_{pa}^{L} \right) \\
&s.t.~\mathbf{M}^{sp}_{c} \in\{0,1\}^{(k-1) \times M}, \\
&\mathbf{M}^{sp \top}_{c} \mathbf{1}^{k-1}=\mathbf{1}^M, ~~\mathbf{M}^{sp}_{c} \mathbf{1}^M=\frac{M}{k-1} \mathbf{1}^{k-1}.
\end{split}
\end{aligned}
\end{small}
\end{equation}
The doubly stochastic nature~\cite{sinkhorn1964relationship} of $\mathbf{\Pi}^{sp}_{c}$ encourages a sparse matching with part-level diversity, which is critical to avoid the trivial solution: all visual nodes are wrongly matched to a single linguistic counterpart (see Table~\ref{tab:analysis}). Thus, each positive entry in $\mathbf{\Pi}^{sp}_{c}$ indicates a matched visual and linguistic node pair across two subgraphs.

Inspired by human cognition to identify unseen class objects through retrieving seen class object characteristics, such as parts, we explicitly mine shared knowledge among all classes and adaptively implement knowledge transfer from seen to unseen classes. To this end, we first select the most matched node in $\mathcal{G}_{pa}^{L}$ for each node of $\mathcal{G}_{sp}^{V}$ in class $c$ according to $\mathbf{\Pi}^{sp}_{c}$, and then retrieve other class object sharing the same part to derive the constraint of $\mathcal{L}_{match}^{sp}$. For instance, as in Figure \ref{fig:F_SPMatch}, the visual feature of the region $m_N$ (i.e. semantic embedding $q_N$) is matched with `cat eye' in the affinity matrix. Thus, we pick out text embeddings of `person eye' and `bird eye' from $\mathcal{G}_{pa}^{L}$ and enforce the semantic embedding $q_N$ close to these picked text embeddings. 
For other object parts occurring in other classes, such as `person finger' and `bird wing', we encourage the semantic embedding $q_N$ to stay away from those text embeddings. 

Formally, we calculate similarity scores $s'(t_{m,c}^{pa}, q'_n)=1/(1+e^{-\frac{1}{\tau} s(t_{m,c}^{pa}, q'_n)})$ between the semantic embedding $q_n$ and all text embeddings in $\mathcal{G}_{pa}^{L}$ (i.e. $\{t_{m, c}^{pa}\}_{m=1, c=1}^{M, C_s+C_u}$), and then constrain the derived scores by the supervision signal of $\mathbf{L}^{sp}=\{0,1\}^{M\times(C_s+C_u)}$, which is defined through the following rules. For each semantic embedding $q_n$ in $\mathcal{G}_{sp}^{V}$, if the text embedding $t_{m, c}^{pa}$ in $\mathcal{G}_{pa}^{L}$ contains the same part information, the label of this text embedding should be 1. Then, if the text embedding $t_{m, c}^{pa}$ in $\mathcal{G}_{pa}^{L}$ contains different objects as well as part information, the label of this text embedding should be 0; otherwise, no supervision signal is generated. Hence, the spatial-part graph matching loss is defined as follows:
\begin{equation}
\mathcal{L}_{match}^{sp} =\sum_{m=1}^{M} \sum_{c=1}^{C_s+C_u}\sum_{n=2}^{k} \left \| s'(t_{m,c}^{pa}, q'_n) - \mathbf{L}_{m, c}^{sp} \right \|^2.
\label{eq:sp_match}
\end{equation}
The minimized loss value ensures dissimilarity between irrelevant object parts, enhancing discriminative visual feature extraction. Moreover, through minimizing $\mathcal{L}_{match}^{sp}$, we can mine shared knowledge in terms of object parts, and semantic relations among classes are explicitly transferred from linguistic to visual representations, facilitating unseen class object recognition in visual feature space. 

\subsection{Channel-State Matching (CSMatch) Module}\label{chVLMatch}

\begin{table*}[tp!] 
\centering
\noindent
\caption{{Comparison with existing GZS3 methods on PASCAL VOC \cite{everingham2015pascal} and COCO-Stuff \cite{caesar2018coco} datasets. The `Seen', `Unseen', and `Harm.' denote mIoU of seen classes, unseen classes, and their harmonic mean. The best results are highlighted in \textbf{bold}.}}
\scalebox{0.9}{\begin{tabular}{c | c | c | c | c | c c c | c c c }
\toprule[1pt]
\xrowht{8pt}
\multirow{2}{*}{Method}&\multirow{2}{*}{Embed}&\multirow{2}{*}{Type}&\multirow{2}{*}{Self-training}&\multirow{2}{*}{Setting}&\multicolumn{3}{c |}{PASCAL VOC}&\multicolumn{3}{c }{COCO-Stuff}\\
\xrowht{8pt}
&&&&&Seen&Unseen&Harm.&Seen&Unseen&Harm.\\
\hline
CaGNet \cite{gu2020context}&Word2vec&{generative}&\XSolidBrush&inductive&78.4&26.6&39.7&33.5&12.2&18.2\\
CaGNet \cite{gu2020context}&Word2vec&{generative}&\CheckmarkBold&transductive&78.6&30.3&43.7&35.6&13.4&19.5\\
SIGN \cite{cheng2021sign}&Word2vec&{generative}&\XSolidBrush&inductive&75.4&28.9&41.7&32.3&15.5&20.9\\
SIGN \cite{cheng2021sign}&Word2vec&{generative}&\CheckmarkBold&transductive&83.5&41.3&55.3&31.9&17.5&22.6\\

\xrowht{6pt}
STRICT \cite{pastore2021closer}&Word2vec&{embedding}&\CheckmarkBold&transductive&82.7&35.6&49.8&35.3&30.3&32.6\\
Joint \cite{baek2021exploiting}&Word2vec&{embedding}&\XSolidBrush&inductive&77.7&32.5&45.9&-&-&-\\


ZSSeg \cite{xu2022simple}&CLIP&{embedding}&\XSolidBrush&inductive&83.5&72.5&77.5&{39.3}&36.3&37.8\\

ZegFormer \cite{ding2022decoupling}&CLIP&{embedding}&\XSolidBrush&inductive&86.4&63.6&73.3&36.6&33.2&34.8\\

ZegFormer+CSL$_{2}$ \cite{zhang2024csl}&CLIP&{embedding}&\XSolidBrush&inductive&87.1&68.6&76.9&37.5&36.2&36.9\\

PADing \cite{he2023primitive}&CLIP&{embedding}&\XSolidBrush&inductive&-&-&-&\textbf{40.4}&24.8&30.7\\
Ours (DeVLMatch)&CLIP&{embedding}&\XSolidBrush&inductive&86.8&71.5&78.4&38.1&36.6&37.3\\

Ours (DeVLMatch)&CLIP&{embedding}&\XSolidBrush&transductive&\textbf{87.0}&\textbf{74.1}&\textbf{80.0}&{38.3}&\textbf{37.7}&\textbf{38.0}\\


\bottomrule[1pt]
\end{tabular}}
\label{Table:cv}
\end{table*}

\subsubsection{Channel visual graph $\mathcal{G}_{ch}^{V}$}We humans usually comprehend an object by identifying states of objects, such as texture and color, rather than recognize only one overview characteristic. To capture the texture and color information \cite{vittayakorn2016automatic, liang2020training} in channels, we decouple objects through grouping channels of the most matched \textit{semantic embedding} $q_n \in Q_{m}, q_n\in \mathbb{R}^{d}$ for each class, and each group $\{q_{n_{r}} \in \mathbb{R}^{\frac{d}{R}}\}_{r=1}^{R}$ encodes one state of the corresponding class object. Then, the derived groups are fed into a linear layer to obtain $\{q''_{n_{r}} \in \mathbb{R}^{d}\}_{r=1}^{R}$, which are regarded as nodes of $\mathcal{G}_{ch}^{V}$ {(Figure \ref{fig:F_CSMatch})}. Within the mini-batch, nodes belonging to the same class are connected. 

\subsubsection{State linguistic graph $\mathcal{G}_{st}^{L}$}To build the state linguistic graph $\mathcal{G}_{st}^{L}$  {(Figure \ref{fig:F_CSMatch})}, we introduce state words defined in \cite{isola2015discovering}, including 115 adjectives $\mathcal{S}=\{S_1, S_2, ..., S_{115}\}$. Hence, we obtain $115\times(C_s+C_u)$ nodes to establish the graph $\mathcal{G}_{st}^{L}$, where each node ${t}_{m, c}^{st}$ is represented as the text embedding of a phrase made by the state $S_{m}$ and object $O_{c}$ pair, such as `fur sheep'. Then, we use ConceptNet to produce relational scores between object states $\mathcal{S}$ and object classes $\mathcal{O}$. Top $M$ states are selected for each class, among which edges are connected inner the same class or same part. 

\subsubsection{Channel-state graph matching}Given the built channel visual graph $\mathcal{G}_{ch}^{V}$ and part linguistic graph $\mathcal{G}_{st}^{L}$, we perform the channel-state graph matching to bridge the gap between visual and linguistic comprehension from channel perspective  {(Figure \ref{fig:F_CSMatch})}. In particular, a node affinity matrix $\mathbf{M}^{cs}_{c}$ is computed to measure the correlation of nodes across visual and linguistic subgraphs for class $c$. Then, a Sinkhorn \cite{sinkhorn1964relationship} is performed to sparse the affinity matrix $\mathbf{M}^{cs}_{c}$ and obtain $\mathbf{\Pi}^{cs}_{c}$, modeling the subgraph matching between $\hat{\mathcal{G}}_{ch}^{V}$ and $\hat{\mathcal{G}}_{st}^{L}$. According to $\mathbf{\Pi}^{cs}_{c}$, we select the most matched node in $\mathcal{G}_{st}^{L}$ for each node of $\mathcal{G}_{ch}^{V}$ and then diffuse the matching across the whole state linguistic graph. Similar to the spatial-part graph matching, by retrieving other class objects sharing the same state, we can derive the supervision signal of $\mathbf{L}^{cs}=\{0,1\}^{M\times(C_s+C_u)}$. Therefore, the channel-state graph matching loss is defined as follows:
\begin{equation}
\mathcal{L}_{match}^{cs} = \sum_{m=1}^{M} \sum_{c=1}^{C_s+C_u} \sum_{r=1}^{R} \left \| s'(t_{m,c}^{st}, q''_{n_{r}}) - \mathbf{L}_{m,c}^{cs} \right \|^2.
\label{eq:cs_match}
\end{equation}
By optimizing $\mathcal{L}_{match}^{cs}$, the shared knowledge in terms of object states across classes is explicitly mined, which can enhance knowledge transfer from seen to unseen classes in visual feature space for benefiting GZS3.

\section{Experiments}
\subsection{Datasets and Implementation}
\textbf{PASCAL-VOC dataset} \cite{everingham2015pascal} provides 1464 training images and 1449 test images with segmentation annotations of 20 object classes. Following \cite{gu2020context, xian2019semantic, ding2022decoupling}, we split the dataset into 15 seen classes and 5 unseen classes.

\textbf{COCO-Stuff dataset} \cite{caesar2018coco} is composed of 118287 training images and 5000 validation images. The corresponding pixel-level annotations contain 171 object classes in total. Following \cite{xian2019semantic, xian2018zero, ding2022decoupling}, we regard 156 classes as seen classes and the rest 15 classes as unseen classes. 

\textbf{CATARACTS dataset} \cite{luengo20212020} contains 4670 images captured from cataract surgery videos with segmentation labels. We split label classes into seen and unseen classes according to their semantic relationships, and thus obtain 6 seen classes (`pupil', `iris', `cornea', `surgical tape', `eye retractors', `hand') and 2 unseen classes (`skin', `instrument').

\textbf{Evaluation Metrics} include the mean intersection-over-union (mIoU, \%) over seen and unseen classes, and their harmonic mean, which are widely used in GZS3 works \cite{ding2022decoupling,xu2022simple,zhang2024csl}. 

\textbf{Implementation Details.} Our implementation is based on Detectron2 \cite{wu2019detectron2}. We use ResNet101 \cite{he2016deep} as our backbone, FPN \cite{lin2017feature} as the pixel decoder, and ViT-B/16 CLIP model \cite{radford2021learning} as the text encoder. Two proposed modules are plug-and-play as well as readily implementable. The number ($N$) and dimension ($d$) of queries are empirically set as 100 and 256 by default. We use a batch size of 32. We use the ADAMW as our optimizer with a learning rate of 0.0001 and 1e-4 weight decay. Hyper parameters $k$, $R$, $\alpha$, $\beta$ are empirically set as 3, 4, 2, 2. 

\subsection{Benchmark Comparison}
\subsubsection{Performance on PASCAL VOC and COCO-Stuff} 
As shown in Table \ref{Table:cv}, we can observe that the proposed DeVLMatch shows superior performance in all evaluation metrics, outperforming other competing methods (w/ ResNet101 as backbone) by a large margin. Specifically, DeVLMatch surpasses SIGN (w/ self-training), the representative generative-based method, with increments of 24.7\% and 15.4\% in terms of harmonic mean score on PASCAL VOC and COCO-Stuff datasets, respectively. Notably, both generative and self-training methods require complex multi-stage training schemes, and they also need to be retrained whenever new unseen classes are incoming. Moreover, the self-training strategy requires access to unlabeled pixels of unseen classes. On the contrary, DeVLMatch is an embedding-based method that can be applied on-the-fly to any unseen classes. Compared to the state-of-the-art embedding-based ZegFormer \cite{ding2022decoupling} method, our DeVLMatch shows superiority, especially in unseen class mIoU, with increments of 10.5\% and 4.5\% on PASCAL VOC and COCO-Stuff, verifying the effectiveness of DeVLMatch in reasoning unseen class through mining shared knowledge with decoupled object information. 

{In addition, we report the performance of our method in the inductive setting, where neither the unseen class names nor the corresponding images are accessible during the training phase. To adhere to this strict condition, we discard the unseen class graph nodes in part and state linguistic graphs (i.e., $\mathcal{G}_{pa}^{L}$ and $\mathcal{G}_{st}^{L}$). Consequently, supervision signals of $\mathbf{L}^{sp}$ and $\mathbf{L}^{cs}$ in Eq. (\ref{eq:sp_match}, \ref{eq:cs_match}) exclude the unseen class. Despite the absence of unseen class information during training in the inductive setting, the fine-grained part and state information facilitate the transfer of knowledge from seen to unseen classes, surpassing the baseline \cite{ding2022decoupling}.}

\begin{table}[tp!] 
\caption{GZS3 performance on medical image segmentation.}
\centering
\noindent
{\begin{tabular}{c | c c c }
\toprule[1pt]
Method&Seen&Unseen&Harmonic\\
\hline
SPNet& 59.2 &5.5& 10.0\\
ZegFormer& 64.8 &11.9& 20.2\\
\cellcolor[gray]{.93} Ours (DeVLMatch)&\cellcolor[gray]{.93} \textbf{69.3}&\cellcolor[gray]{.93} \textbf{15.8}&\cellcolor[gray]{.93} \textbf{25.8}\\
\bottomrule[1pt]
\end{tabular}}
\label{Table:medical}
\end{table}

\begin{figure*}[t]
\centering
\includegraphics[width=172mm]{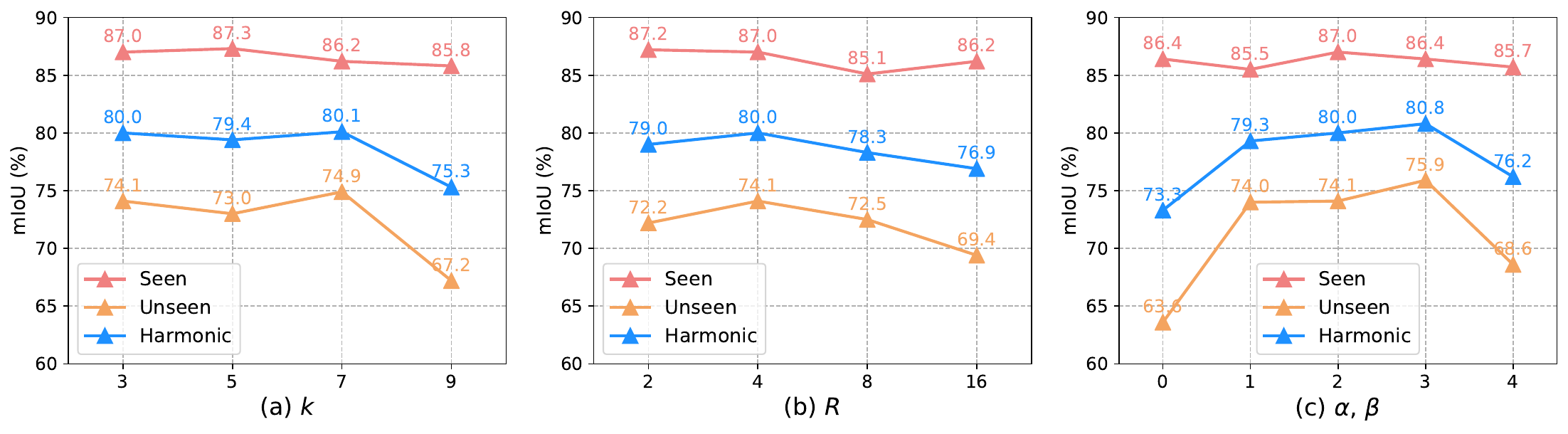}
\caption{Sensitivity analysis on hyper parameters, including spatial query number $k$, channel split number $R$, weighting coefficients $\alpha$, $\beta$.}
\label{fig:hyper}
\end{figure*}

\subsubsection{Performance on Medical Image Segmentation} Since we are the first to explore GZS3 task in medical image segmentation, we have no existing methods to compare against. We compare DeVLMatch to SPNet \cite{xian2019semantic} and ZegFormer \cite{ding2022decoupling}, which are implemented with the same codebase and common settings with our DeVLMatch for a fair comparison. As compared in Table \ref{Table:medical}, it is obvious that our method significantly surpasses \cite{xian2019semantic}, \cite{ding2022decoupling}  by 15.8\% and 5.6\% harmonic mean score of mIoU, demonstrating DeVLMatch can be well adapted to medical image segmentation scenario. 

\subsection{Analytical Results}

\subsubsection{Ablation study}To investigate the effectiveness of individual components in DeVLMatch, we conduct ablation experiments on PASCAL VOC and COCO-Stuff datasets, and comparison results are shown in Table \ref{Table:ablation}. We first evaluate the proposed two components in DeVLMatch independently. In particular, SPMatch and CSMatch modules individually outperform the baseline model \cite{ding2022decoupling} with increments of 3.9\%, 2.1\% in harmonic mean value. Then, we can observe that the complete version of DeVLMatch gives the highest performance on all datasets, 
revealing SPMatch and CSMatch modules are mutually complementary.

\begin{table}[tp!] 
\caption{Ablation study on PASCAL VOC and COCO-Stuff.}
\centering
\noindent
{\begin{tabular}{p{1.85cm} | p{0.5cm} p{0.8cm} p{0.7cm}  | p{0.5cm} p{0.8cm} p{0.7cm} }
\toprule[1pt]
\xrowht{8pt}
\multirow{2}{*}{Method}&\multicolumn{3}{c|}{PASCAL VOC}&\multicolumn{3}{c }{COCO-Stuff}\\
\xrowht{8pt}
&Seen&Unseen&Harm.&Seen&Unseen&Harm.\\
\hline
Baseline&86.4&63.6&73.3&36.6&33.2&34.8\\
w/ SPMatch&86.1&70.0&77.2 &38.0 &37.2&37.6\\
w/ CSMatch&85.0&67.7&75.4 & 38.2 &35.2 &36.6\\
\cellcolor[gray]{.93} DeVLMatch&\cellcolor[gray]{.93} \textbf{87.0}&\cellcolor[gray]{.93} \textbf{74.1}&\cellcolor[gray]{.93} \textbf{80.0}&\cellcolor[gray]{.93} \textbf{38.3}&\cellcolor[gray]{.93} \textbf{37.7}&\cellcolor[gray]{.93} \textbf{38.0}\\
\bottomrule[1pt]
\end{tabular}}
\label{Table:ablation}
\end{table}

\subsubsection{Sensitivity analysis on hyper parameters $k$, $R$, $\alpha$, $\beta$} 

\textit{{1) Spatial query number $k$:}} In SPMatch, top $k$ similar \textit{semantic embeddings} are selected to represent object parts for class $c$ in visual space. Herein, we set up different spatial query number $k$ in the process of $\mathcal{G}_{sp}^{V}$ construction, and results on PASCAL VOC are shown in Figure \ref{fig:hyper} (a). It is clear that the performance of our model remains stable when $k<=7$. A large $k$ value ($k$=9) results in degraded performance. This may be because spatial regions belonging to other classes are selected to indicate object parts of the current class, leading to unreasonable spatial visual graph construction and biased graph matching.

\textit{{2) Channel split number $R$:}} In CSMatch, $\mathcal{G}_{ch}^{V}$ is constructed by splitting channels of semantic embedding $q_n \in \mathbb{R}^{d}$ into $R$ groups, and each group encodes one kind of object state. We investigate the effect of channel split number $R$ in Figure \ref{fig:hyper} (b). We can observe that a low $R$ value ($R$=2) cannot sufficiently decouple object and encode all state information, showing 1.0\% decrement in harmonic mean. Though splitting channels into more groups can alleviate the problem, each group may be unable to represent one object state information, degrading performance when $R$=16. Therefore, we choose $R$=4 to reach a trade-off. 

\textit{{3) Weighting coefficients $\alpha$ and $\beta$:}} The hyper parameters $\alpha$ and $\beta$ dominate the contribution of spatial-level $\mathcal{L}_{match}^{sp}$ and channel-level $\mathcal{L}_{match}^{cs}$. To study the influence of SPMatch and CSMatch modules, we vary $\alpha$, $\beta$ values in the range of 0 to 4 with a step size of 1. Note that the model with $\alpha$, $\beta$ equal to 0 is our baseline, ZegFormer \cite{ding2022decoupling}. It is observed that introducing SPMatch and CSMatch benefits the GZS3 task and leads to 10.5\% mIoU improvement in unseen classes when the value of $\alpha$, $\beta$ is set as 3.

\subsubsection{Correspondences among $\mathcal{G}_{sp}^{V}$, $\mathcal{G}_{ch}^{V}$, $\mathcal{G}_{pa}^{L}$, $\mathcal{G}_{st}^{L}$}To provide further insight into the proposed DeVLMatch, we investigate the correspondences among constructed four graphs, including $\mathcal{G}_{sp}^{V}$, $\mathcal{G}_{ch}^{V}$, $\mathcal{G}_{pa}^{L}$, and $\mathcal{G}_{st}^{L}$. In original implementation, based on human intuition, $\mathcal{G}_{sp}^{V}$ is designed to match with $\mathcal{G}_{pa}^{L}$ in SPMatch module, and $\mathcal{G}_{ch}^{V}$ is matched to $\mathcal{G}_{st}^{L}$ in CSMatch module. Herein, we create three additional variants ($3^{rd}$-$5^{th}$ rows) to analyze the correspondences. As shown in Table \ref{Table:4graphs}, we can observe that all variants are superior to the baseline ($2^{nd}$ row), demonstrating the effectiveness of the decoupling strategy to comprehend objects. Compared with three variants, our matching pair design of DeVLMatch ($6^{th}$ row) performs best. Moreover, completely swapping the matching pairs ($3^{rd}$ row) results in the worst performance among three variants. These realities reveal that the working mechanism of GZS3 model conforms to human intuition, justifying our pairing design.

\begin{table}[tp!] 
\caption{Correspondence of $\mathcal{G}_{sp}^{V}$, $\mathcal{G}_{pa}^{L}$, $\mathcal{G}_{ch}^{V}$, $\mathcal{G}_{st}^{L}$ on PASCAL VOC.}
\centering
\noindent
{\begin{tabular}{p{0.75cm} |  p{0.75cm}  | p{0.75cm}  | p{0.75cm}  | c c c }
\toprule[1pt]
\xrowht{8pt}$\mathcal{G}_{sp}^{V} \Leftrightarrow \mathcal{G}_{pa}^{L}$&$\mathcal{G}_{sp}^{V} \Leftrightarrow \mathcal{G}_{st}^{L}$&$\mathcal{G}_{ch}^{V} \Leftrightarrow \mathcal{G}_{pa}^{L}$&$\mathcal{G}_{ch}^{V} \Leftrightarrow \mathcal{G}_{st}^{L}$&Seen&Unseen&Harm.\\
\hline
&&&&86.4&63.6&73.3\\
&\CheckmarkBold&\CheckmarkBold&&85.6 & 69.3 & 76.6\\
\CheckmarkBold&&\CheckmarkBold&&84.8 & 73.9 &79.0\\ 
&\CheckmarkBold&&\CheckmarkBold& 86.8 &72.1&78.8\\
\cellcolor[gray]{.93} \CheckmarkBold&\cellcolor[gray]{.93} &\cellcolor[gray]{.93} &\cellcolor[gray]{.93} \CheckmarkBold&\cellcolor[gray]{.93} \textbf{87.0}&\cellcolor[gray]{.93} \textbf{74.1}&\cellcolor[gray]{.93} \textbf{80.0}\\
\bottomrule[1pt]
\end{tabular}}
\label{Table:4graphs}
\end{table}

\subsubsection{Sensitivity analysis on quality of $\mathcal{G}_{pa}^{L}$, $\mathcal{G}_{st}^{L}$} When constructing $\mathcal{G}_{pa}^{L}$ and $\mathcal{G}_{st}^{L}$, top $M$ object part and state pairs are selected for each class with respect to relational scores derived from ConceptNet \cite{speer2017conceptnet}. Herein, we reduce quality of $\mathcal{G}_{pa}^{L}$ and $\mathcal{G}_{st}^{L}$ to analyze the sensitivity, as in Table  \ref{tab:senanalysis}. Specifically, we make nodes uniform for each class by only choosing top 1 object part and state pairs (2$^{nd}$ row), and then randomly sample $M$ nodes from top $2M$ pairs (3$^{rd}$ row). Uniform graphs result in unsatisfactory performance, as object cannot be explicitly decoupled and relationships among classes are hard to mine. With random graphs, our DeVLMatch can still boost the performance of Baseline \cite{ding2022decoupling} significantly, demonstrating the robustness to some low-quality nodes of linguistic graphs. The reasons are in two aspects: (1) Our graph matching strategies find out the most matched object part or state node for each visual feature, avoiding undesirable node matching cross graphs. (2) The inter-class relations within linguistic graphs can benefit the knowledge transfer from seen to unseen classes, which are more crucial than the individual node information. Thus, even if some nodes may be counterintuitive, leading to low-quality graphs, the performance could still be enhanced.

\begin{table}[t!]
    \centering
		\caption{Sensitivity analysis on quality of $\mathcal{G}_{pa}^{L}$, $\mathcal{G}_{st}^{L}$  on PASCAL VOC.}
  {\begin{tabular}{c | c c c }
		\toprule[1pt]
		\xrowht{8pt}
		Method&Seen&Unseen&Harm.\\
		\hline
		Uniform&85.1&66.4&74.6\\
		Random&85.2&72.6&78.4\\
		\cellcolor[gray]{.93} ConceptNet \cite{speer2017conceptnet}&\cellcolor[gray]{.93} \textbf{87.0}&\cellcolor[gray]{.93} \textbf{74.1}&\cellcolor[gray]{.93} \textbf{80.0}\\
		\bottomrule[1pt]
		\end{tabular}}
    \label{tab:senanalysis}
\end{table}

\begin{table}[t!]
    \centering
    \caption{Effectiveness of Sinkhorn on PASCAL VOC.}
     {\begin{tabular}{c | c c c }
		\toprule[1pt]
		\xrowht{8pt}
		Method&Seen&Unseen&Harm.\\
		\hline
		Baseline&86.4&63.6&73.3\\
		w/o Sinkhorn&85.1&66.3&74.5\\
		\cellcolor[gray]{.93} Ours (w/ Sinkhorn)&\cellcolor[gray]{.93} \textbf{87.0}&\cellcolor[gray]{.93} \textbf{74.1}&\cellcolor[gray]{.93} \textbf{80.0}\\
		\bottomrule[1pt]
		\end{tabular}}
    \label{tab:analysis}
\end{table}

\subsubsection{Effectiveness of Sinkhorn}In DeVLMatch, we introduce Sinkhorn to establish the spatial-part and channel-state graph matching. To study the effectiveness of Sinkhorn, we ablate it, leading to a decrement of 7.8\% in unseen class mIoU, as in Table \ref{tab:analysis}. This is because that without Sinkhorn, all nodes within the visual graph may be matched with a single linguistic counterpart. This trivial solution leads to insufficient representation for each class and limits the shared knowledge mining among classes, thereby significantly degrading performance in unseen classes.


\subsubsection{Computational efficiency}\label{eff}

\begin{table}[t!]
\centering
\caption{The comparison of computational efficiency between the proposed DeVLMatch framework and baseline ZegFormer.}
{\begin{tabular}{c  c  c  c  c}
\toprule[1pt]
\multirow{2}{*}{Method}&Train time&Test time&\multirow{2}{*}{Harm.}\\
&(s / iter)&(s / iter)&\\
\hline
ZegFormer&1.289&0.177&73.3\\
\cellcolor[gray]{.93} DeVLMatch&\cellcolor[gray]{.93} 1.593&\cellcolor[gray]{.93} 0.177&\cellcolor[gray]{.93} 80.0\\
\bottomrule[1pt]
\end{tabular}}
\label{efficiency}
\end{table}

We adopt ZegFormer \cite{ding2022decoupling} as our backbone and further advance spatial-part matching (SPMatch) module along with channel-state matching (CSMatch) module to bridge the gap between visual and linguistic comprehension for GZS3. Two proposed modules are plug-and-play as well as parameter-free. Compared to \cite{ding2022decoupling}, the training time increases by 23.6\%, as shown in Table \ref{efficiency}. We adopt the same inference process as \cite{ding2022decoupling} for a fair comparison. Under the same inference speed, the proposed DeVLMatch framework achieves a performance gain of 9.1\%, demonstrating the effectiveness of two proposed modules in bridging the comprehension gap between vision and language during the training stage.

\begin{figure}[t]
\centering
\includegraphics[width=88mm]{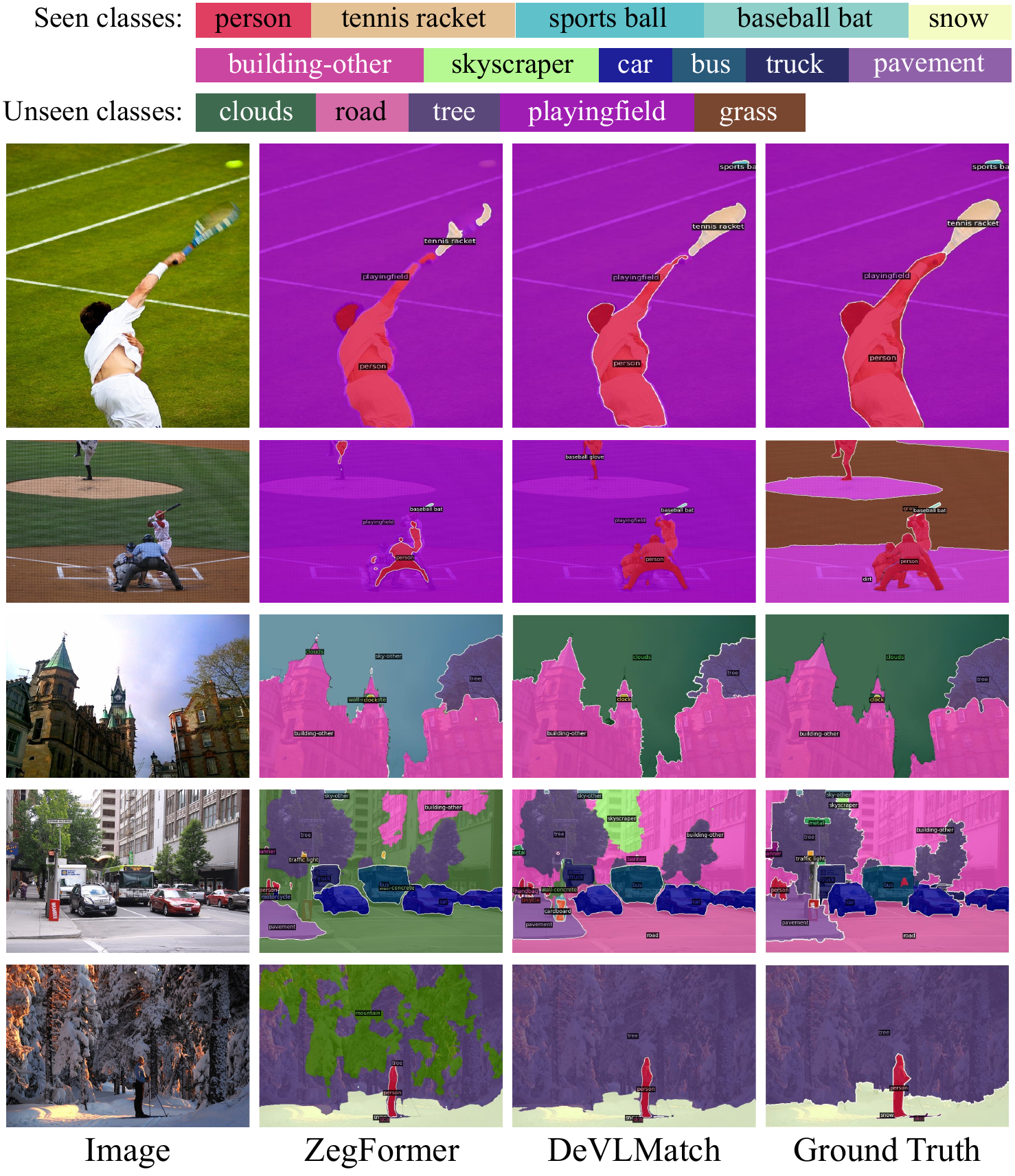}
\caption{Segmentation results on COCO-Stuff dataset, using 171 class names in COCO-Stuff to generate text embeddings.}
\label{fig:seg_vis}
\end{figure}

\subsection{Visualization Results}

\subsubsection{Segmentation results}In qualitative aspect, compared to the baseline ZegFormer, DeVLMatch is superior in both \textbf{\textit{mask}} and \textbf{\textit{class}} predictions. As in $1^{st}$-$2^{nd}$ rows of Figure \ref{fig:seg_vis}, we observe that baseline only activates most representative regions of objects (e.g., `person', `tennis racket', `sports ball'). This is because with a single semantic representation to associate each class, the one-to-one matching leads to inefficient training issues \cite{jia2022detrs}. Owing to our decoupling strategy that can comprehensively represent each class and enable fine-grained object comprehension, DeVLMatch is able to capture object part local features. Hence, in inference, integrating $N$ \textit{class-agnostic segments}, including local part segments, our method can rectify some incomplete \textbf{\textit{mask}} predictions of baseline. Moreover, SPMatch and CSMatch modules can facilitate the alignment between visual features and linguistic semantics in training stage via decoupling and matching objects across graphs. Thus, the linguistic graph involving seen and unseen class relations can better act as a bridge to enhance the generalizability from seen to unseen classes in visual space, facilitating the unseen \textbf{\textit{class}} predictions, as shown in $3^{th}$-$5^{th}$ rows.  

\subsubsection{Segmentation results on extended test classes}\label{extend}

\begin{figure}[t]
\centering
\includegraphics[width=88mm]{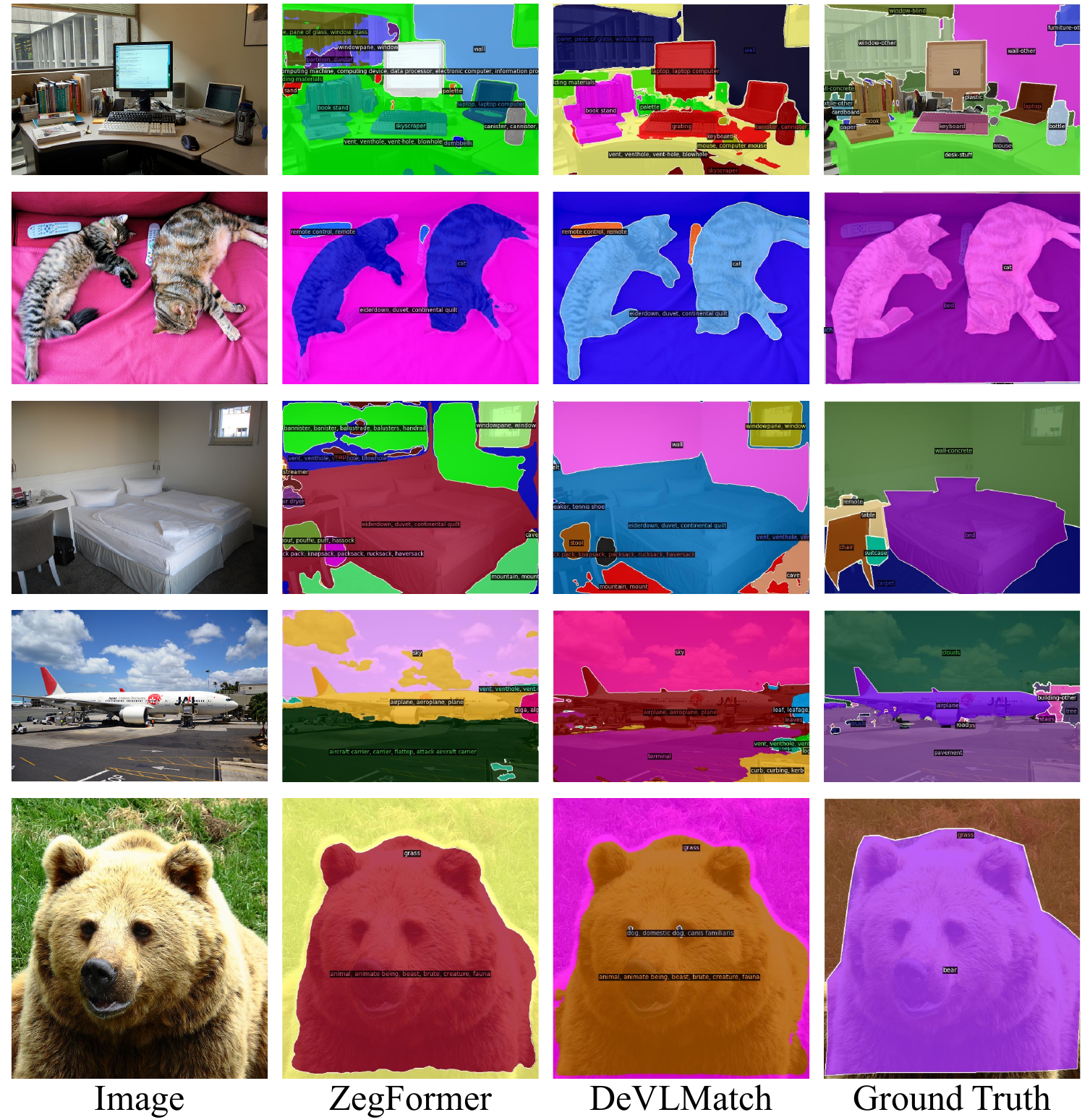}
\caption{Segmentation results on COCO-Stuff dataset, utilizing 847 class names in ADE20k-Full \cite{zhou2017scene} to generate text embeddings.}
\label{fig:seg_extend}
\end{figure}

\begin{figure*}[t]
\centering
\includegraphics[width=1.0\linewidth]{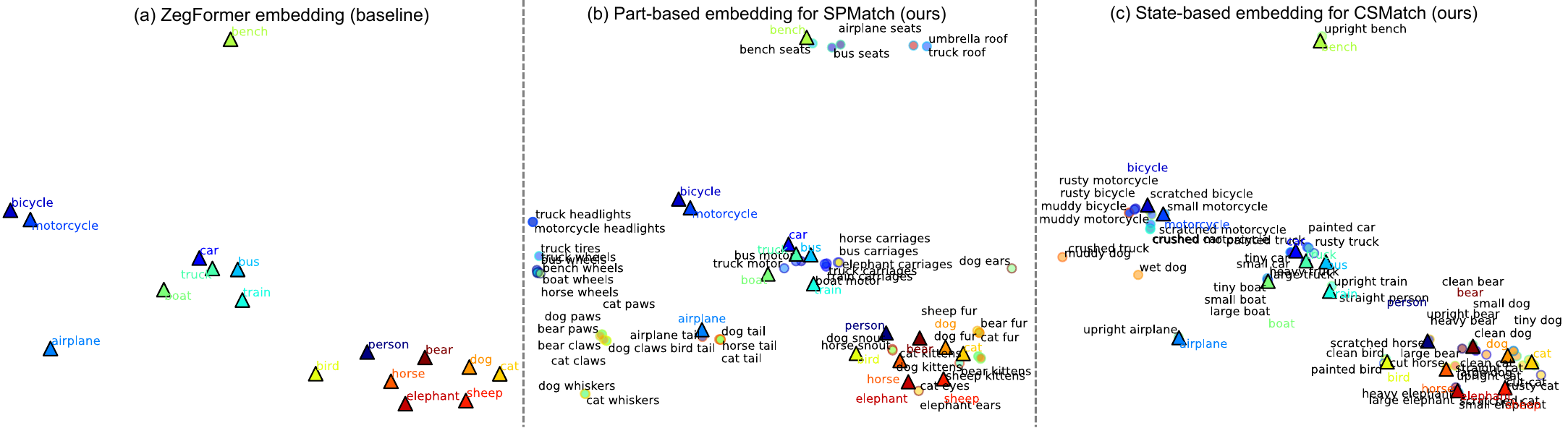}
\caption{Visualization of text embedding. (a) ZegFormer  \cite{ding2022decoupling}  embedding. (b) Part-based embedding for SPMatch. (c) State-based embedding for CSMatch.}
\label{fig:textembedding}
\end{figure*}

\begin{figure*}[h!]
\centering
\includegraphics[width=1.0\linewidth]{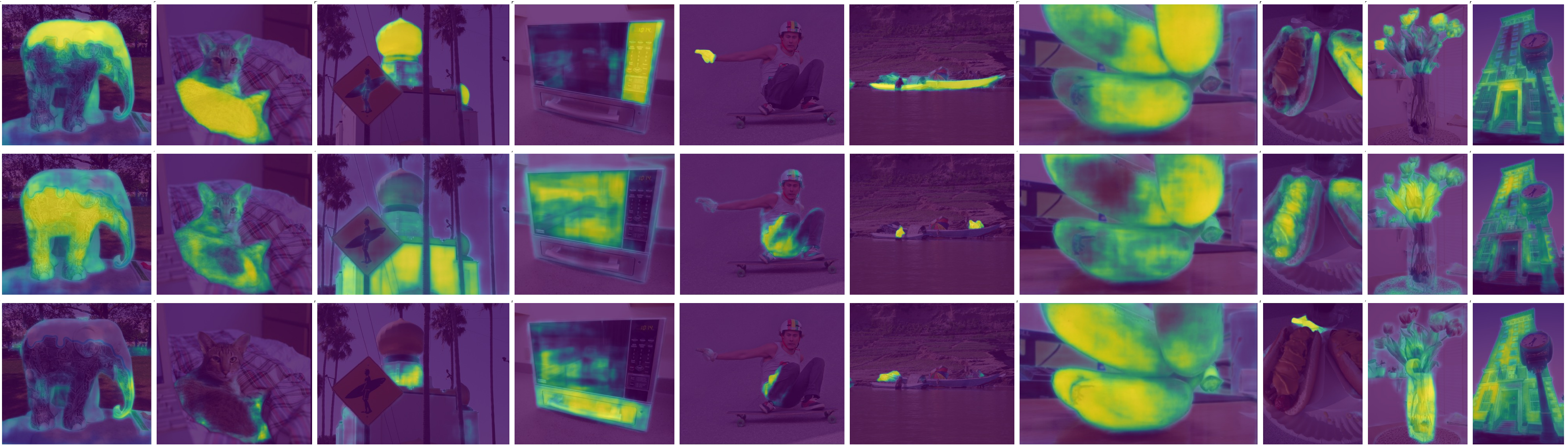}
\caption{{Illustration of the learned parts from object queries. Only with class-level labels, our method is able to learn fine-grained part-level semantics via graph matching.}}
\label{fig:part_vis}
\end{figure*}

The proposed DeVLMatch framework is an embedding-based method that can be applied on-the-fly to any unseen class. To demonstrate this property, we train DeVLMatch with 156 seen classes on {COCO-Stuff dataset} \cite{caesar2018coco}, and then test the the optimized model with extended sets of class names. Specifically, during test phase, we conduct the inference process on COCO-Stuff test data but use 847 class names from ADE20k-Full \cite{zhou2017scene}. The segmentation results are shown in Figure \ref{fig:seg_extend}. Note that the illustrated ground truths are from the COCO-Stuff test dataset with label space containing 156 seen classes and 15 unseen classes. 

Compared to the state-of-the-art embedding-based ZegFormer \cite{ding2022decoupling} method, our DeVLMatch is demonstrated to be superior. For example, different from the baseline model cannot segment the whole region of target object, our method can correctly recognize object regions, such as `computer mouse' in $1^{st}$ row, `remote' and `cat' in $2^{nd}$ row, `wall' in $3^{rd}$ row of Figure \ref{fig:seg_extend}.  Predictions of baseline may not match the environment, such as `alga' in $4^{th}$ row, while our method segments the corresponding region as `leaf', which is more precise. 

The extended set of class names contains richer information to generate diverse text embeddings for inference. Under this circumstance, the proposed DeVLMatch derive some interesting yet reasonable predictions. For example, in $4^{th}$ row, our method recognizes the `pavement' area as `terminal', which is consistent with our common sense. In $5^{th}$ row, DeVLMatch is able to identify bear eye and body regions, revealing the differentiated feature representations among different object parts. This result demonstrates the effectiveness of proposed DeVLMatch framework in decoupling object parts.


\begin{figure*}[t]
\centering
\includegraphics[width=1.0\linewidth]{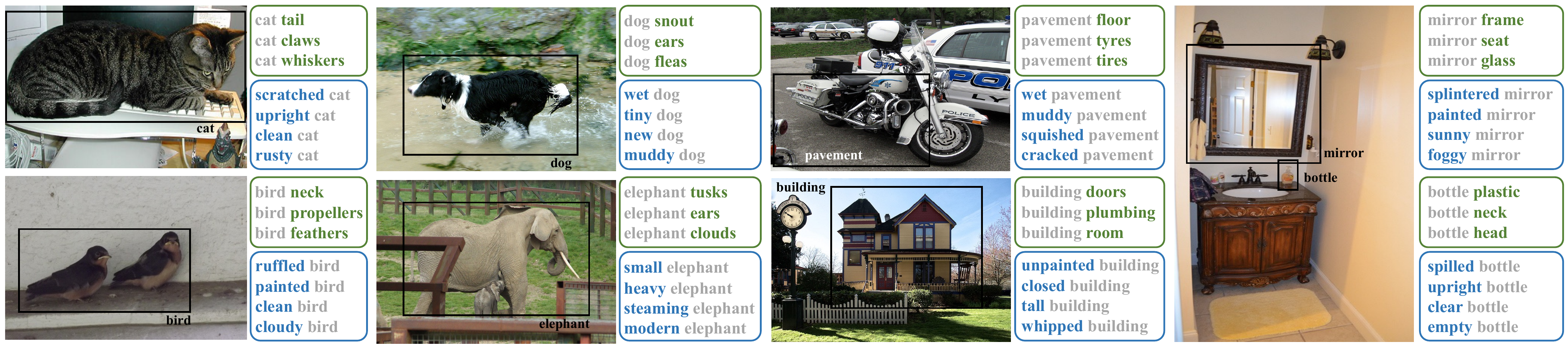}
\caption{{Graph matching results on COCO-Stuff dataset.}}
\label{fig:gm_vis1}
\end{figure*}

\subsubsection{Visualization of text embedding in SPMatch and CSMatch} We employ t-SNE \cite{maaten2008visualizing} to visualize the text embedding of ZegFormer alongside our SPMatch and CSMatch modules, in order to expound on the working mechanism of our proposed method. As depicted in Figure \ref{fig:textembedding}, we observe that the embeddings for class names such as 'truck' and 'motorcycle' exhibit a substantial separation, indicating a dissimilar semantic relationship. However, in reality, 'truck' and 'motorcycle' should share some degree of semantic similarity in specific object parts. Furthermore, within the SPMatch module, we can discern that the embeddings for 'truck headlights' and 'motorcycle headlights' are situated close to each other, signifying a high level of semantic similarity at a fine-grained level. This explicit inter-class relationship facilitates knowledge transfer between classes, ultimately benefiting the GZS3 task, particularly in the context of unseen classes. Within the CSMatch module, we observe that the introduction of `state' for each class leads to a form of label perturbation. As illustrated in prior works \cite{muller2019does, liu2023model}, label perturbation techniques, such as label smoothing \cite{szegedy2016rethinking}, have been demonstrated to prevent overfitting and enhance model calibration for both in-distribution and out-of-distribution data. In light of this, our proposed CSMatch module may contribute to enhancing the generalizability of unseen classes through label perturbation.

\subsubsection{{Visualization of learned object parts from semantic embeddings}} 

{In Figure \ref{fig:part_vis}, we illustrate the representation of learned object parts through \textit{semantic embeddings} $q_n \in Q_{um}$, each corresponding to a local region $m_n$. We can observe the learned \textit{semantic embeddings} represent diverse object parts independently, revealing the success of decoupling for objects in visual space. This is because, with the proposed graph matching methods, the label perturbation induced by incorporating `part' and `state' for each class in linguistic graphs explicitly enhances fine-grained object attribute learning in visual space, further facilitating the relationship mining among classes.}

\begin{figure}[t]
\centering
\includegraphics[width=90mm]{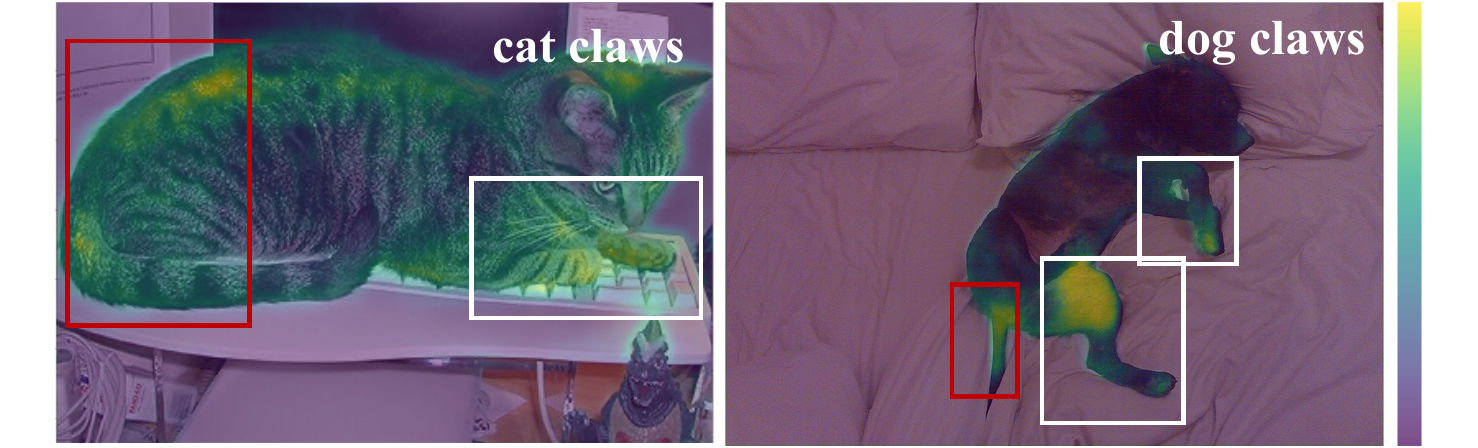}
\caption{{Visualization of the region matched with part of `claws' on COCO-Stuff dataset.}}
\label{fig:gm_vis2}
\end{figure}


\subsubsection{Spatial-part and channel-state graph matching results}
{To intuitively understand the proposed DeVLMatch, we visualize the matched part and state information of objects, as shown in Figure \ref{fig:gm_vis1}. We can observe that many matching results are consistent with human intuition, verifying our method could decouple and bridge the gap between visual and linguistic comprehension. For example, the object `cat' is matched with the part of `cat tail' and the state of `scratched cat'. Despite some unreasonable matching problems shown in the figure, such as `elephant clouds', our method can mine shared part and state knowledge among classes (e.g., `dog ears' and `elephant ears') with the guidance of linguistic cues, which is important for the knowledge transfer from seen to unseen classes.}

In Figure \ref{fig:gm_vis2}, we then visualize the selected region ($m_{n}$) in $\mathcal{G}_{sp}^{V}$, which are matched with the part of `claws' according to the proposed spatial-part graph matching. It is observed that some regions are indeed `claws' (white box), while others are incorrect (red box). Even though, both red boxes highlight the part of `tail', indicating our method can mine shared knowledge among classes with the guidance of linguistic cues, facilitating the knowledge transfer from seen to unseen classes.


\subsection{{Extention of DeVLMatch}}

\subsubsection{{Extention of DeVLMatch to new backbone ZegCLIP}}
{Owing to the play-and-plug nature, our method can be applied on the state-pf-the-art backbone ZegCLIP \cite{zhou2023zegclip}. Different from ZegFormer which performs cross-modality alignment on object tokens, ZegCLIP implements Text-Patch Matching on patch tokens. Hence, we plug the proposed DeVLMatch before the Text-Patch Matching decoder of ZegCLIP in the training phase. In the visual graphs $\mathcal{G}_{sp}^{V}$ and $\mathcal{G}_{ch}^{V}$, graph nodes are derived from patch tokens, while the graph edges are built by linking the patch tokens within the same class. We compare our method with the backbone ZegCLIP in both inductive and transductive settings. As shown in Table \ref{Table:ZegCLP}, the proposed DeVLMatch can further boost the performance of ZegCLIP.}

\begin{table}[t!] 
\centering
\noindent
\caption{{Extention of DeVLMatch to new backbone ZegCLIP \cite{zhou2023zegclip} on COCO-Stuff dataset.}}
\begin{tabular}{p{1.8cm}| p{1.9cm}| p{.7cm} p{1.1cm} p{1.1cm} }
\toprule[1pt]
Method &Setting &Seen & Unseen &Harm. \\\hline
ZegCLIP&Inductive & 40.2	 & 41.1	 & 40.8 \\
DeVLMatch &Inductive &\textbf{41.5}	 & \textbf{43.4}	 & \textbf{42.4} \\\hline
ZegCLIP &Transductive & 40.7	&59.9	&48.5\\
DeVLMatch &Transductive & \textbf{41.3}	&\textbf{61.3} 	&\textbf{49.6}          \\
\bottomrule[1pt]
\end{tabular}
\label{Table:ZegCLP}
\end{table}

\begin{table}[t!] 
\centering
\noindent
\caption{{Extention of DeVLMatch by
introducing SAM \cite{kirillov2023segment} on COCO-Stuff.}}
\begin{tabular}{p{4.7cm}| p{.7cm} p{.85cm} p{.85cm}}
\toprule[1pt]
Method&Seen & Unseen &Harm. \\\hline
DeVLMatch&{41.3}&{61.3} &{49.6} \\
DeVLMatch w/ SAM (sub-parts)&{41.8}&{61.4}&{49.7}          \\
DeVLMatch  w/ SAM (parts) &\textbf{42.1}&\textbf{62.0}&\textbf{50.2}          \\
\bottomrule[1pt]
\end{tabular}
\label{Table:sam}
\end{table}

\subsubsection{{Extention of DeVLMatch by introducing SAM}}

{To construct spatial visual graph $\mathcal{G}_{sp}^{V}$, we compute similarities between matched \textit{semantic embeddings} $Q_m$ and unmatched counterparts $Q_{um}$ within the training batch and then select top k similar \textit{semantic embeddings} in $Q_{um}$ per class to represent object parts in $\mathcal{G}_{sp}^{V}$. Herein, we explore SAM \cite{kirillov2023segment} to detect object parts and build $\mathcal{G}_{sp}^{V}$ in our SPMatch module, and we conduct this experiment upon the backbone ZegCLIP.} 

\begin{figure}[t!]
\centering
\includegraphics[width=1.0\linewidth]{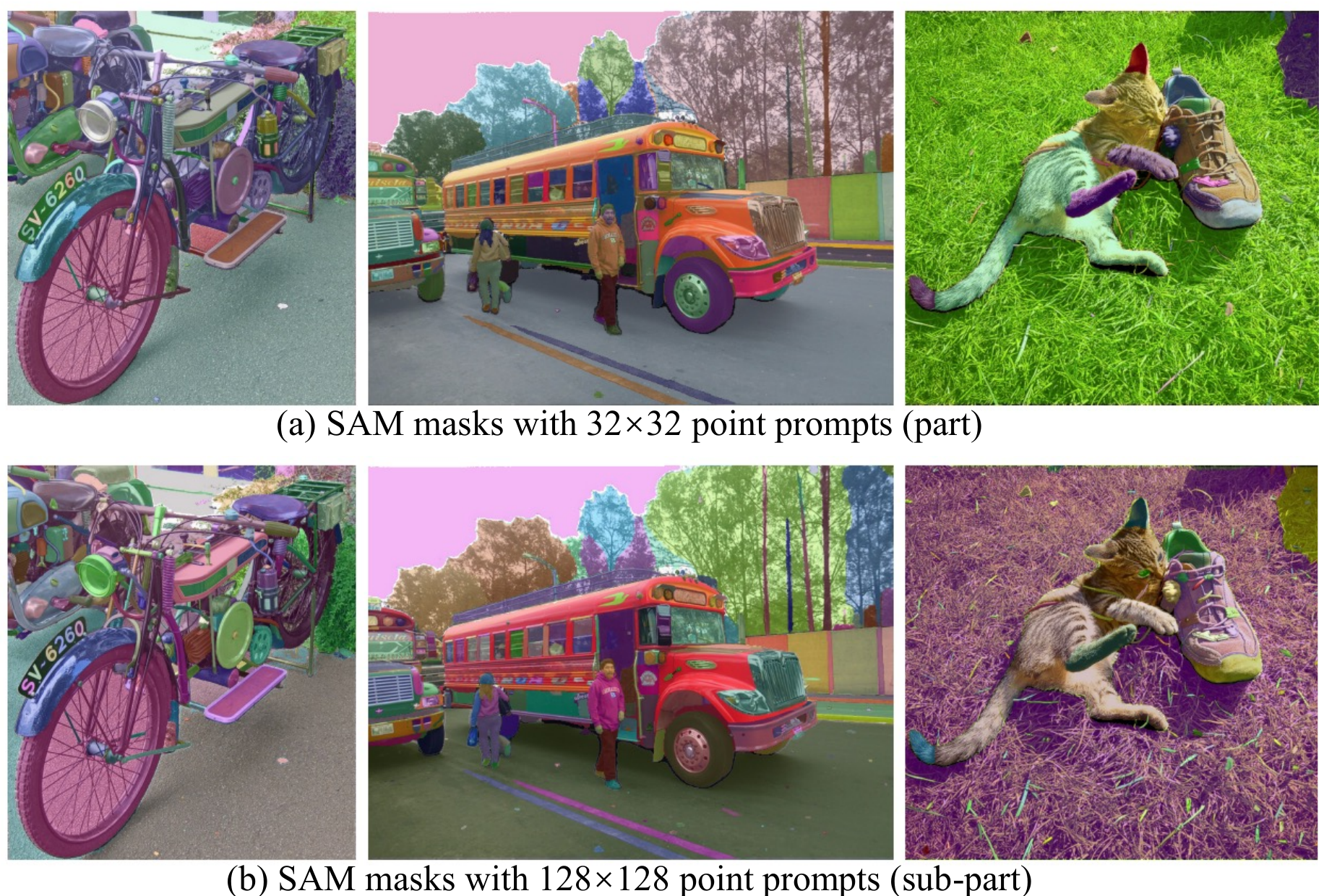}
\caption{{Illustration of the generated SAM mask with point-level prompts in different granularity: (a) 32$\times$32 and (b) 128$\times$128 point prompts.}}
\label{fig:sam_mask}
\end{figure}

{We first use SAM with point-level prompts to generate masks in different granularity to explicitly model parts and sub-parts of objects, as shown in Figure~\ref{fig:sam_mask}. Compared with 32$\times$32 prompts, using more fine-grained prompts (e.g., 128$\times$128) can generate sub-parts, such as the tree in the second image. For each input training image $x$, $N$ \textit{patch embeddings} are obtained from the CLIP image encoder, and we have segmentation ground truth $y\in \mathbb{Z}^{H\times W}$ with $C_s$ seen classes. Through SAM, $K$ masks can be derived for the input image, and the average \textit{patch embedding} within each mask represents a graph node in $\mathcal{G}_{sp}^{V}$. As for the graph edge, we leverage the semantic-level constraint \cite{jiang2022graph, zhang2021prototypical} by linking the queries within the same class. To find out graph nodes within the same class, we calculate the pair-wise IoU between segmentation ground truth per class and $K$ masks derived from SAM. If one SAM mask shows the largest IoU with the segmentation ground truth of class $c$, the corresponding graph node belongs to class $c$. Then the following operators, including constructing $\mathcal{G}_{pa}^{L}$ and performing spatial-part graph matching, keep the same as our previous implementation in SPMatch module. We follow the inference process in ZegCLIP \cite{zhou2023zegclip} for a fair comparison.}

{The comparison results are shown in Table \ref{Table:sam}. We can observe the introduction of SAM refines the quality of graph nodes in $\mathcal{G}_{sp}^{V}$, leading to more accurate cross-modality alignment at a fine-grained level. Notably, aligning for sub-parts yields inferior performance compared to aligning for object parts. This discrepancy may be attributed to the presence of meaningless sub-parts, such as individual alphabets in the first example of Figure \ref{fig:sam_mask} within the response letter.}

\section{Conclusion}
In this paper, we propose a DeVLMatch framework for GZS3 task, which decouples and matches objects across visual and linguistic perspectives. The decoupling strategy is inspired by the observation that humans often use some detailed `part' and `state' information to comprehend seen objects and recognize unseen classes. Graph matching is implemented to bridge the gap between visual and linguistic comprehension in decoupled object part and state levels, thereby explicitly introspecting class relationships among seen and unseen classes. Based on this insight, we devise spatial-part SPMatch and channel-state CSMatch modules. The former decouples and matches objects with spatial part information from both visual and linguistic perspectives. The latter matches object states in the linguistic aspect to compatible channel information in the visual aspect. Extensive experimental results demonstrate significant advantages of DeVLMatch over the previous GZS3 works. 




\end{document}